\documentclass[lettersize,journal]{IEEEtran}
\usepackage{amsmath,amsfonts}
\usepackage{amsthm}
\usepackage{algorithm}
\usepackage{algorithmic}
\usepackage{multirow}
\usepackage{multicol}
\usepackage{tabularx}
\usepackage{enumitem}
\usepackage{array}
\usepackage[caption=false,font=normalsize,labelfont=rm,textfont=rm]{subfig}
\usepackage{textcomp}
\usepackage{stfloats}
\usepackage{url}
\usepackage{verbatim}
\usepackage{booktabs}
\usepackage{graphicx}
\usepackage{cite}
\usepackage[font=small]{caption}
\usepackage{hyperref}
\usepackage{color}
\usepackage[dvipsnames]{xcolor}
\usepackage{xcolor,colortbl}
\definecolor{tabGray}{gray}{0.94}

\theoremstyle{remark}
\newtheorem{remark}{\bf Remark}

\begin{document}

\title{Robust State Estimation for Legged Robots \\
with Dual Beta Kalman Filter}

\author{Tianyi Zhang, Wenhan Cao, Chang Liu, Tao Zhang, Jiangtao Li, Shengbo Eben Li
\thanks{This study is supported by Tsinghua University-EFORT Intelligent Equipment Co., Ltd Joint Research Center for Embodied Intelligence Computing and Perception.}
\thanks{Tianyi Zhang is with the School of Vehicle and Mobility, Tsinghua University, Beijing, China (e-mail: zhangtia24@mails.tsinghua.edu.cn).}
\thanks{Wenhan Cao is with the State Key Laboratory of Intelligent Green Vehicle and Mobility, Tsinghua University, Beijing, China (e-mail: cwh19@mails.tsinghua.edu.cn).}
\thanks{Chang Liu is with the Department of Advanced Manufacturing
and Robotics, College of Engineering, Peking University, Beijing, China (e-mail: changliucoe@pku.edu.cn).
}
\thanks{T. Zhang and J. Li are with SunRising AI Ltd as chief research scientists. They both graduated from School of Vehicle and Mobility, Tsinghua University, China (e-mail: zhang.t1983@gmail.com and  andy.ljt1988@gmail.com).}
\thanks{Shengbo Eben Li is with the School of Vehicle and Mobility and College
of Artificial Intelligence, Tsinghua University, Beijing, China (e-mail: lish04@gmail.com).}

\thanks{Corresponding Author: Shengbo Eben Li}
}

\markboth{Journal of \LaTeX\ Class Files,~Vol.~14, No.~8, August~2021}%
{Shell \MakeLowercase{\textit{et al.}}: A Sample Article Using IEEEtran.cls for IEEE Journals}


\maketitle

\begin{abstract}
Existing state estimation algorithms for legged robots that rely on proprioceptive sensors often overlook foot slippage and leg deformation in the physical world, leading to large estimation errors. To address this limitation, we propose a comprehensive measurement model that accounts for both foot slippage and variable leg length by analyzing the relative motion between foot contact points and the robot’s body center. We show that leg length is an observable quantity, meaning that its value can be explicitly inferred by designing an auxiliary filter. To this end, we introduce a dual estimation framework that iteratively employs a parameter filter to estimate the leg length parameters and a state filter to estimate the robot’s state. 
To prevent error accumulation in this iterative framework, we construct a partial measurement model for the parameter filter using the leg static equation. This approach ensures that leg length estimation relies solely on joint torques and foot contact forces, avoiding the influence of state estimation errors on the parameter estimation. Unlike leg length which can be directly estimated, foot slippage cannot be measured directly with the current sensor configuration. However, since foot slippage occurs at a low frequency, it can be treated as outliers in the measurement data. 
To mitigate the impact of these outliers, we propose the $\beta$-Kalman filter ($\beta$-KF), which redefines the estimation loss in canonical Kalman filtering using $\beta$-divergence. This divergence can assign low weights to outliers in an adaptive manner,  thereby enhancing the robustness of the estimation algorithm. These techniques together form the dual $\beta$-Kalman filter (Dual $\beta$-KF), a novel algorithm for robust state estimation in legged robots. Experimental results on the Unitree GO2 robot demonstrate that the Dual $\beta$-KF significantly outperforms state-of-the-art methods.

\end{abstract}


\section{Introduction}
\label{sec.intro}
\IEEEPARstart{L}{egged} robots have garnered increasing attention in recent years \cite{bledt2018cheetah, bloesch2013state, bloesch2013state2}. Their leg actuators provide greater degrees of freedom compared to the wheels of traditional robots \cite{fink2020proprioceptive}, enabling them to perform more complex tasks, such as navigating rough terrains, overcoming obstacles, and operating in unstructured environments \cite{bledt2018cheetah}. Achieving these capabilities relies on accurately estimating the robot’s state, including its pose and velocity \cite{bloesch2013state}. As a result, various algorithms have been developed \cite{bloesch2013state,bloesch2013state2,fink2020proprioceptive,yang2023multi,yoon2023invariant, teng2021legged} to achieve accurate state estimation of legged robots from noisy measurements.

\begin{figure}[!t]
\centering
\begin{minipage}{0.49\linewidth}
    \centering
    \subfloat[]{\includegraphics[width=0.95\textwidth,height=1.3\textwidth]{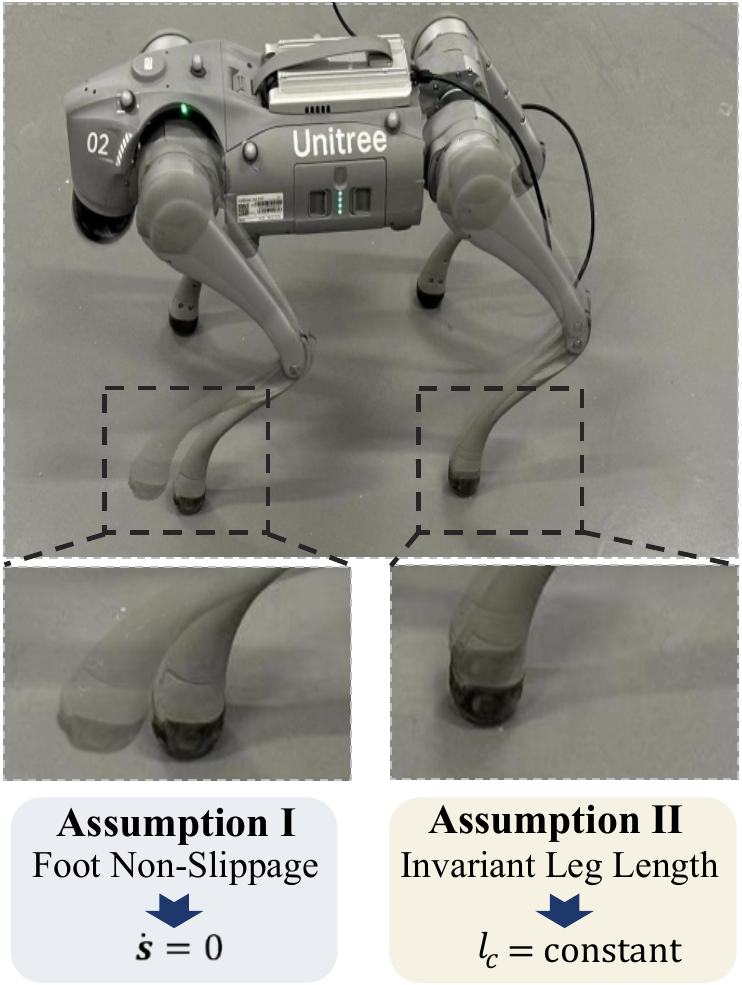}
    \label{fig1.a}}
\end{minipage}
\hfill
\begin{minipage}{0.49\linewidth}
    \centering
    \includegraphics[width=1\textwidth,height=0.62\textwidth]{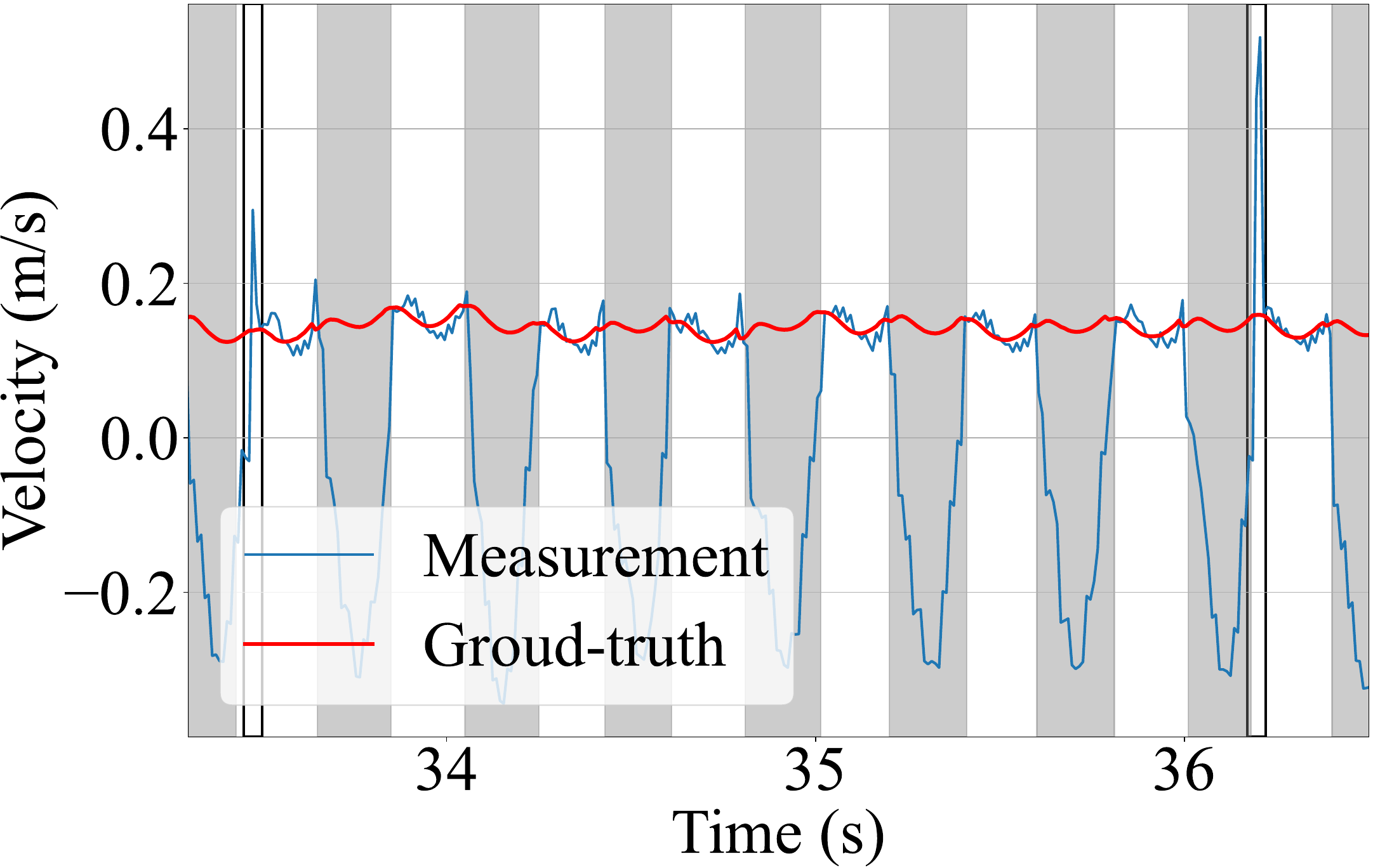}
    \\
    
    \subfloat[]{\includegraphics[width=1\textwidth,height=0.63\textwidth]{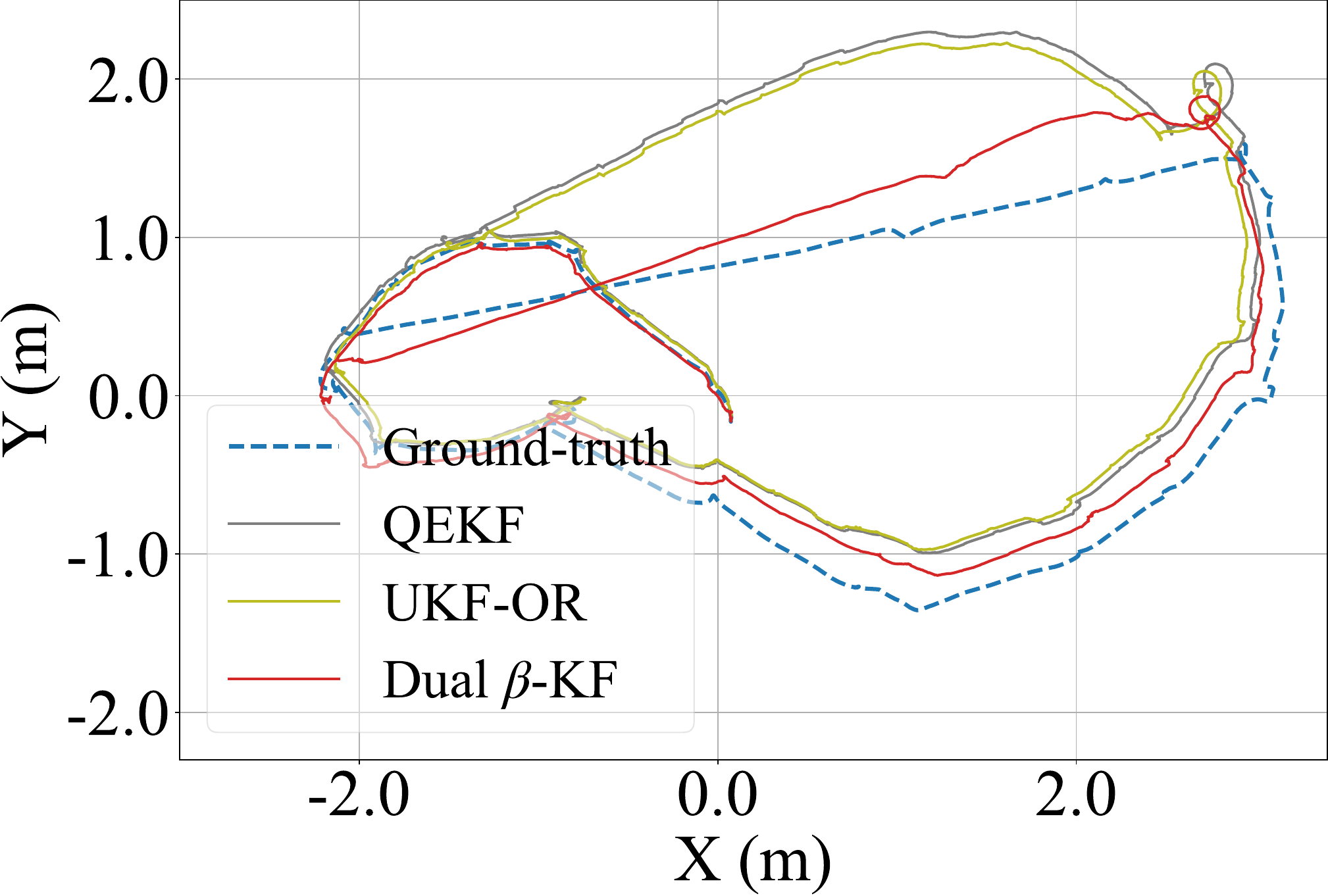}\label{fig1.c}}
\end{minipage}
\caption{(a) Foot slippage and dynamic deformation occur during the locomotion of legged robots. (b) Above: Comparison of the body velocity calculated using the existing measurement model \cite{bloesch2013state2} with the true velocity obtained from the motion capture system. The regions marked by thin black bars indicate significant spikes. The grey-shaded areas mark periods when the leg is off the ground, during which body velocity measurement is meaningless \cite{yang2022online}. Below: Estimated trajectories from different algorithms, with our proposed Dual $\beta$-KF being the closest to the ground-truth trajectory.}
\label{fig1}
\end{figure}

Existing state estimation algorithms for legged robots rely on two types of sensors: proprioceptive and exteroceptive sensors. The proprioceptive sensors measure the internal condition of the robot and include devices such as inertial measurement units (IMUs), joint encoders, and torque/force sensors \cite{bloesch2013state, yang2023multi, yoon2023invariant}. 
In contrast, the exteroceptive sensors gather information from the robot's external environment, consisting of sensors like cameras and lidar to help the robot perceive its surroundings \cite{teng2021legged, wisth2022vilens}.
Generally speaking, leveraging the rich information from exteroceptive sensors enhances state estimation accuracy \cite{teng2021legged, wisth2022vilens}. However, these sensors are often costly and introduce significant computational overhead due to the complexity of processing large data volumes \cite{wisth2022vilens, yoon2023invariant}. Additionally, in extreme weather conditions like heavy rain, snow, or fog, lidar and camera measurements can become unreliable due to signal scattering \cite{yoon2023invariant}. Therefore, for low-cost legged robots, it is highly beneficial to achieve accurate state estimation using only proprioceptive sensors, which are generally more affordable, computationally efficient and trustworthy.

Proprioceptive state estimation algorithms for legged robots rely on two critical models: the state propagation model and the measurement model. The state propagation model describes the kinematic evolution of the robot’s state, including the body's pose and velocity, as well as the position of each foot. In this model, the body's acceleration and angular velocity serve as control inputs, whose values can be obtained from IMU. The measurement model typically characterizes the residual error between two representations of the relative foot kinematics, i.e., the position and velocity of the foot contact point relative to the body center \cite{bloesch2013state, bloesch2013state2, fink2020proprioceptive}. In detail, the first representation is the vector difference between the position and velocity of the body and foot contact points, while the second employs forward kinematics (FK) to calculate the relative position and velocity based on joint angles and velocities.

Unlike the state propagation model that can be easily established through IMU dynamics, constructing the measurement model in previous works relies on two significant assumptions: (i) the contacting foot is assumed to have zero velocity, and (ii) the leg length is assumed to be constant (see Fig.\ref{fig1.a}). 
Unfortunately, these assumptions are often violated due to foot slippage \cite{bloesch2013state2, kim2021legged, camurri2017probabilistic} and dynamic leg deformation \cite{bloesch2013kinematic, yang2022online, wisth2022vilens} during locomotion. As illustrated in Fig.\ref{fig1.c}, we calculate the body velocity using the measurement model from prior studies \cite{bloesch2013state2, yang2023multi} and compare it to the actual body velocity obtained from a motion capture system. The results reveal irregular spikes in the estimated values, indicating that the violation of these assumptions prevents the model from accurately capturing the true physical characteristics, which can introduce large estimation errors if we directly use this misspecified measurement model for state estimation \cite{cao2022robust}.

To eliminate these two assumptions in the measurement model, we incorporate foot slippage velocity into the first representation of the relative position between foot and body center, and explicitly account for leg length variations in the second representation. However, this well-considered measurement model introduces two new challenges: foot slippage velocity and leg length values cannot be directly measured. A simple observability analysis reveals that foot slippage velocity can be regarded as structured unknown information, meaning its actual values cannot be measured with the current sensor configuration. 
In contrast, leg length can be viewed as parameterized unknown information, whose values are observable by using foot contact forces and joint torques. Building on this analysis, we develop a novel state estimation framework for our proposed measurement model called dual beta Kalman filter, or Dual $\beta$-KF for short. This framework enables the estimation of leg length through a dual estimation approach while mitigating the effects of foot slippage by constructing a more robust filtering algorithm with a modified objective function. Experimental results demonstrate that this framework achieves higher estimation accuracy compared to widely used filtering algorithms, such as quaternion-based extended Kalman filter (QEKF) \cite{bloesch2013state, fink2020proprioceptive} and unscented Kalman filter with outlier rejection (UKF-OR) \cite{bloesch2013state2}. The specific contributions of this paper can be summarized as follows:

\begin{itemize}
\item{We demonstrate that the assumptions of foot non-slippage and constant leg length in existing measurement models are not valid in practical scenarios, leading to significant estimation errors if applied directly. To address this issue, we develop a comprehensive measurement model that explicitly accounts for foot slippage and leg length by analyzing the relative motion between the legged robot's foot contact points and its body center, enabling a more accurate physical representation of the robot.} 

\item {For the purpose of accurately acquiring leg length parameters, we propose a dual estimation framework that simultaneously estimates the leg length using a parameter filter and the robot’s state using a state filter. To prevent the accumulation of errors in this dual structure, we construct a leg static measurement model for the parameter filter. This kind of partial model allows us to estimate leg length solely through joint torques and foot contact forces, without relying on state information, thereby ensuring that state information does not interfere with parameter estimation.} 

\item {Accurately measuring foot slippage velocity remains challenging with existing sensor configurations. However, due to the low frequency of foot slippage, it can be regarded as outliers of the measurement data. To mitigate the impact of these outliers, we design a robust state filter inspired by an optimization perspective of conventional Kalman filters. Specifically, we reinterpret the Kalman filter as a solution to a maximum a posteriori (MAP) estimation problem based on Kullback-Leibler (KL) divergence. By redefining the estimation loss using a more robust $\beta$-divergence, we develop a novel filtering approach called the $\beta$-Kalman filter ($\beta$-KF), which reduces the influence of foot slippage on estimation accuracy.} 

\item{Combined the dual estimation framework with the robust $\beta$-KF, we propose a new algorithm for robust state estimation for legged robots called dual beta Kalman filter. Experiments on both Gazebo simulation and a real-world Unitree GO2 robot demonstrate that our proposed method significantly improves estimation accuracy compared to state-of-the-art proprioceptive state estimation algorithms.}
\end{itemize}

The remainder of the paper is organized as follows: Section \ref{sec:related} provides a review of the previous research in the field. Section \ref{sec:problem} constructs the state-space model for the legged robot state estimation problem. Section \ref{sec:method} details the proposed algorithms. Section \ref{sec:exp} presents the results in both Gazebo simulation and real-world experiments. 

\section{Related Works}
\label{sec:related}
In this section, we discuss related legged robot state estimation algorithms using only proprioceptive sensors that address scenarios where the two assumptions introduced in Sec.\ref{sec.intro} are not met, i.e. foot slippage and varying leg length. The algorithms can be broadly categorized into two groups: slip detection to suppress foot slippage-induced outliers, and kinematic calibration to obtain accurate leg length.

\noindent\textbf{Slip detection: } 
The earliest and most common approaches for slip detection involve algorithms that use handcrafted thresholds \cite{bloesch2013state2, kim2021legged} to detect outliers in the measurement data. If the threshold is exceeded, the measurement will be considered an outlier and discarded to maintain estimation accuracy. Building on this foundation, Bloesch et al \cite{bloesch2013state2} first proposed an UKF-based proprioceptive sensors fusion algorithm, which sets a certain threshold for the Mahalanobis distance of the filter innovation to detect slippage. Later, Kim et al. \cite{kim2021legged} improved this algorithm by setting a threshold for the \( \ell_2 \)-norm of foot velocity, rather than filter innovation. Although these threshold-based algorithms have been applied to various estimators to enhance robustness \cite{fink2020proprioceptive,teng2021legged,yoon2023invariant}, they lead to loss of the measurement information near the threshold.

Compared to threshold-based algorithms, recent studies have used machine learning-based approaches to infer slip probability, which is then used to weight and fuse measurement information. For example, Camurri et al. \cite{camurri2017probabilistic} introduced a logistic regressor to estimate the probability of reliable contact for each foot, i.e., to identify slippage. Further, Lin et al. \cite{lin2021legged} replaced the logistic regressor with a more powerful deep neural network, using sensory data as input to predict the probability of stable contact. In addition, many other learning-based approaches employ techniques like clustering \cite{rotella2018unsupervised} and kernel density estimation \cite{maravgakis2023probabilistic} to detect slippage. With weighted fusion, all measurement information is utilized to varying degrees, allowing these learning-based approaches to avoid information loss.
However, they require offline training for different robots and environments, which limits their generalizability. Conversely, our proposed $\beta$-KF is an online optimization algorithm that automatically assigns low weights to outliers, avoiding offline training and preventing information loss.

\noindent\textbf{Kinematics calibration: }Apart from foot slippage, several studies \cite{bloesch2013kinematic, yang2022online} have shown that inaccurate leg length parameters can also significantly affect the estimation accuracy. Based on this fact, Bloesch et al. \cite{bloesch2013kinematic} proposed a kinematic calibration framework to address this issue by formulating calibration as a maximum likelihood estimation problem. This algorithm achieves accurate parameter estimates, but it can only run offline.
Recently, Yang et al. \cite{yang2022online} implemented an online kinematic calibration algorithm that uses an extended Kalman filter (EKF) to estimate both the state and leg length parameters simultaneously, effectively reducing state estimation errors. This algorithm is similar to our proposed dual estimation framework. Nevertheless, it relies on a motion capture system to provide true body velocity to perform the EKF update step, which is impractical in a majority of real-world scenarios. In contrast, our proposed dual filter uses only proprioceptive sensors, eliminating the need for additional external equipment.

\section{State-Space Model Construction for Estimation Problem}
\label{sec:problem}
State estimation for a robotic system generally requires constructing a stochastic state-space model (SSM) \eqref{model}. Typically, this model consists of two parts: a state propagation model \eqref{model.state} and a measurement model \eqref{model.meas}. The former describes the evolution of the legged robot's motion, while the latter explains how sensors measure the state. The SSM is formulated as
\begin{subequations}
\label{model}
\begin{align}
\boldsymbol{x}_{t} &= f(\boldsymbol{x}_{t-1}, \boldsymbol{u}_{t-1}) + \boldsymbol{n}_{t-1} \label{model.state}, 
\\ 
\boldsymbol{y}_{t} &= h(\boldsymbol{x}_{t};\boldsymbol{\varrho}_t) + \boldsymbol{\eta}_t,\label{model.meas}
\end{align}
\end{subequations}
where $f(\cdot, \cdot)$ is the transition function, and $h(\cdot; \cdot)$ is the measurement function; $\boldsymbol{x}_t$ and $\boldsymbol{u}_{t}$ represent the robot's state and control input, respectively; $\boldsymbol{y}_t$ is the measurement vector; $\boldsymbol{\varrho}_t$ denotes the  leg length parameters; and $\boldsymbol{n}_{t}$ and $\boldsymbol{\eta}_t$ represent the propagation and measurement noise, respectively. 
\begin{figure}[!t]
\centering
\includegraphics[width=0.36\textwidth, height=0.2\textwidth]{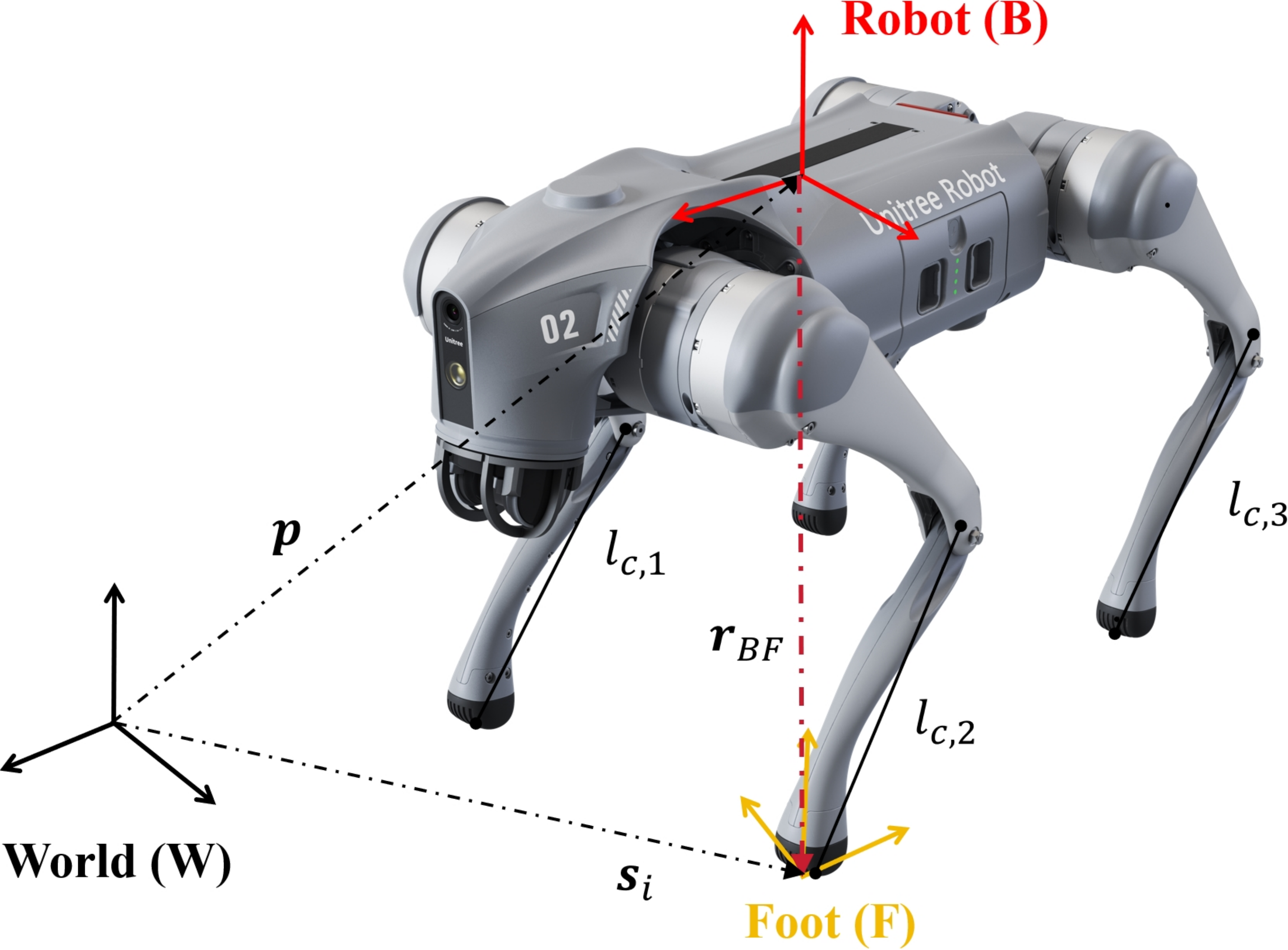}
\caption{Important frame definitions and leg length parameters of Unitree GO2 robot.}
\label{fig_2}
\end{figure}

\subsection{State propagation Model} 
The key aspect of constructing the state propagation model is defining the state of the legged robot. To provide a precise characterization of the motion, we define the full state as
$\boldsymbol{x}:=\left[\boldsymbol{p}; \boldsymbol{v}; \boldsymbol{q}; \boldsymbol{s}_1;...;\boldsymbol{s}_N; \boldsymbol{b}_{\omega}; \boldsymbol{b}_{a}\right],
$ where $\boldsymbol{p} \in \mathbb{R}^3$ and $\boldsymbol{v} \in \mathbb{R}^3$ represent the robot's position and linear velocity in the world frame;  $\boldsymbol{q} \in \mathbb{R}^4$ is the robot’s orientation quaternion; $\{\boldsymbol{s}_i\}_{i=1}^N$ denotes the foot positions in the world frame, where $N$ is the number of legs (e.g., $N=4$ for a quadruped robot); and $\boldsymbol{b}_{\omega}$ and $\boldsymbol{b}_{a}$ are the biases of the IMU gyroscope and accelerometer.

We assume that the IMU frame coincides with the body frame, then the IMU directly measures body angular velocity $\tilde{\boldsymbol{\omega}}$ and linear acceleration $\tilde{\boldsymbol{a}}$ in the body frame:
\begin{equation*}
\label{imu_model}
\tilde{\boldsymbol{\omega}} = \boldsymbol{\omega} + \boldsymbol{b}_{\omega} + \boldsymbol{n}_{\omega}, \ \tilde{\boldsymbol{a}} = \boldsymbol{a} + \boldsymbol{b}_{a} + \boldsymbol{n}_{a}, 
\end{equation*}
where $\boldsymbol{n}_{\omega}$, $\boldsymbol{n}_{a}$ are Gaussian white noise terms, $\boldsymbol{\omega}$ and $\boldsymbol{a}$ represent the true angular velocity and linear acceleration, and the IMU biases $\boldsymbol{b}_{\omega}$, $\boldsymbol{b}_{a}$ are modeled as Brownian motions: 
\begin{equation}
\label{eq.imu_bias}
\dot {\boldsymbol{b}}_{\omega} = \boldsymbol{n}_{b\omega}, \ 
\dot {\boldsymbol{b}}_{a} = \boldsymbol{n}_{ba},
\end{equation}
with $\boldsymbol{n}_{b\omega}$ and $\boldsymbol{n}_{ba}$ representing  Gaussian white noise.

Then the motion equations of the robot \cite{bloesch2013state2, yang2022online} can be derived as  
\begin{equation}
\label{eq.continuous}
\begin{split}
\dot {\boldsymbol{p}} &= \boldsymbol{v}, \quad
\dot {\boldsymbol{v}} = \mathbf{R}(\boldsymbol{q})(\tilde{\boldsymbol{a}} -\boldsymbol{b}_{a} -\boldsymbol{n}_{a}) + \boldsymbol{g}, \\
\dot {\boldsymbol{q}} &= \frac{1}{2}\boldsymbol{q} \otimes \boldsymbol(\tilde{\boldsymbol{\omega}} - \boldsymbol{b}_{\omega} - \boldsymbol{n}_{\omega}),
\end{split}
\end{equation}
where $\boldsymbol{g}$ is the gravitational acceleration, and $\otimes$ denotes quaternion multiplication.  $\mathbf{R}(\boldsymbol{q}) \in SO(3)$ is the rotation matrix in terms of the quaternion $\boldsymbol{q}$.

When the foot is in contact with the ground, the foot position in the body frame can be modeled as a random walk process with  Gaussian white noise $\boldsymbol{n}_s$ \cite{bloesch2013state}, then we have
\begin{equation}
\label{eq.foot}
\dot {\boldsymbol{s}}_i = \mathbf{R}(\boldsymbol{q})\boldsymbol{n}_s.
\end{equation}
After discretizing equations \eqref{eq.imu_bias}, \eqref{eq.continuous}, and \eqref{eq.foot}, we derive the state propagation model \eqref{model.state} as
\begin{equation}
\small
\label{process}
\begin{split}
\begin{bmatrix} 
\boldsymbol{p}_{t} \\ \boldsymbol{v}_{t} \\
\boldsymbol{q}_{t} \\ \boldsymbol{s}_{1,t} \\
...\\
\boldsymbol{s}_{N,t} \\
\boldsymbol{b}_{\omega,t} \\ \boldsymbol{b}_{a,t} 
\end{bmatrix} = \begin{bmatrix} 
\boldsymbol{p}_{t-1} + \boldsymbol{v}_{t-1} \delta t\\ 
\boldsymbol{v}_{t-1} + (\mathbf{R}(\boldsymbol{q}_{t-1})(\tilde{\boldsymbol{a}}_{t-1} -\boldsymbol{b}_{a,t-1}) + \boldsymbol{g}) \delta t\\
\boldsymbol{q}_{t-1} \otimes \boldsymbol{\Phi}((\tilde{\boldsymbol{\omega}}_{t-1} - \boldsymbol{b}_{\omega,t-1}) \delta t) \\ 
\boldsymbol{s}_{1,t-1} \\
...\\
\boldsymbol{s}_{N,t-1} \\
\boldsymbol{b}_{\omega,t-1} \\ 
\boldsymbol{b}_{a,t-1} 
\end{bmatrix} + \boldsymbol{n}_{t-1},
\end{split}
\end{equation}
where $\boldsymbol{\Phi}(\cdot)$ is a map from rotation vector to quaternion \cite{bloesch2013state}, $\delta t$ is the time interval between $t-1$ and $t$. The vector on the right-hand side of \eqref{process} corresponds to $f(\boldsymbol{x}_{t-1}, \boldsymbol{u}_{t-1})$ in \eqref{model.state}, where $\boldsymbol{u}_t \stackrel{\triangle}{=} \left[\tilde{\boldsymbol{a}}_t;\tilde{\boldsymbol{\omega}}_t\right]$. Finally, $\boldsymbol{n}_t$ represents the $(3N + 16)$-dimensional propagation noise.

\subsection{Measurement Model}
\label{sec.measurement}
Following the previous work \cite{bloesch2013state, bloesch2013state2, fink2020proprioceptive}, the measurement model is established using the residuals from two representations of the foot contact point's position and velocity relative to the body.  Of these two representations, one is calculated using forward kinematics based on joint encoder measurements, while the other relies solely on the differences between state variables. In this residual setup, the measurement vector $\boldsymbol{y}_t$ is naturally set to $\boldsymbol{0}$ \cite{bloesch2013state}.

For clarity, we denote $\tilde {\boldsymbol{\phi}}_i \in \mathbb{R}^3$ and $\tilde {\dot {\boldsymbol{\phi}}}_i \in \mathbb{R}^3$ as the $i$-th leg's joint angles and angular velocities measurements  corrupted by white Gaussian noise $\boldsymbol{\eta}_{\phi}$,   $\boldsymbol{\eta}_{\dot \phi}$ as $\tilde {\boldsymbol{\phi}}_i = \boldsymbol{\phi}_i + \boldsymbol{\eta}_{\phi}$, $\tilde {\dot {\boldsymbol{\phi}}}_i = \dot {\boldsymbol{\phi}}_i + \boldsymbol{\eta}_{\dot \phi}$.
These two measurements are provided by the joint encoders.

The relative position between the foot contact point of the 
$i$-th leg and the body center in the world frame \cite{bloesch2013state} can be determined as
\begin{equation}
\label{eq.fk}
\boldsymbol{r}^{\text{BF}}_{i,t} = \mathbf{R}(\boldsymbol{q}_t)\boldsymbol{fk}(\boldsymbol{\phi}_{i,t};\boldsymbol{\varrho}_t).
\end{equation}
This representation is based on the leg forward kinematics function $\boldsymbol{fk}(\cdot;\cdot)$ and joint encoders measurements. Based on the position relationship between the foot contact point and the body center (see Fig.\ref{fig_2}), the vector $\boldsymbol{r}^{\text{BF}}_{i,t}$ can also be expressed solely in terms of state variables as
\begin{equation}
\label{eq.fk_states}
\boldsymbol{r}^{\text{BF}}_{i,t} = \boldsymbol{s}_{i,t}-\boldsymbol{p}_t.
\end{equation}
By taking the derivative of \eqref{eq.fk} and \eqref{eq.fk_states}, we can also obtain two representations for the relative velocity as 
\begin{equation}
\label{eq.BV_meas}
\begin{aligned}
\dot{\boldsymbol{r}}^{\text{BF}}_{i,t} &=\mathbf{R}(\boldsymbol{q}_t)\big(\boldsymbol{J}(\boldsymbol{\phi}_{i,t}; \boldsymbol{\varrho}_t)\dot {\boldsymbol{\phi}}_{i,t} + \left[\boldsymbol{\omega}_{t}\right]_{\times}\boldsymbol{fk}(\boldsymbol{\phi}_{i,t}; \boldsymbol{\varrho}_t)\big) \\
&= \dot {\boldsymbol{s}}_{i,t} -\boldsymbol{v}_t,
\end{aligned}
\end{equation}
where $\left[\boldsymbol{\omega}_t\right]_{\times}$ is the skew-symmetric matrix of $\boldsymbol{\omega}_t$, and $\boldsymbol{J}(\cdot;\cdot)$ is the kinematics Jacobian matrix. The first representation in \eqref{eq.BV_meas} is also referred to as leg odometry (LO) velocity. When foot slippage does not occur, i.e., $\dot {\boldsymbol{s}}_{i,t}=0$, LO velocity serves as the body velocity measurement. The specific expressions for $\boldsymbol{fk}(\cdot;\cdot)$ and $\boldsymbol{J}(\cdot;\cdot)$ are consistent with those in \cite{yang2022online}. By calculating the residual between the two representations of $\boldsymbol{r}^{\text{BF}}_{i,t}$ and $\dot{\boldsymbol{r}}^{\text{BF}}_{i,t}$,
the measurement function $h(\boldsymbol{x}_t;\boldsymbol{\varrho}_t)$ in  \eqref{model.meas} can be defined as
\begin{equation}
\small
\label{obs}
\begin{aligned}
&h(\boldsymbol{x}_t;\boldsymbol{\varrho}_t)  \\
=&\begin{bmatrix}
\{\mathbf{R}(\boldsymbol{q}_t)\boldsymbol{fk}(\tilde{\boldsymbol{\phi}}_{i,t};\boldsymbol{\varrho}_t) -  (\boldsymbol{s}_{i,t} - \boldsymbol{p}_{t})\}_{i=1}^N \\ 
\{\mathbf{R}(\boldsymbol{q}_t)\big(\boldsymbol{J}(\tilde{\boldsymbol{\phi}}_{i,t}; \boldsymbol{\varrho}_t)\tilde{\dot {\boldsymbol{\phi}}}_{i,t} + \left[\tilde{\boldsymbol{\omega}}_{t}\right]_{\times}\boldsymbol{fk}(\tilde{\boldsymbol{\phi}}_{i,t}; \boldsymbol{\varrho}_t)\big) 
- (\dot{\boldsymbol{s}}_{i,t} -\boldsymbol{v}_{t})\}_{i=1}^N 
\end{bmatrix},
\end{aligned}
\end{equation}
and the measurement noise $\boldsymbol{\eta}_t$ is a $6N$-dimensional vector. 
\begin{remark}
In constructing the measurement model, unlike previous works \cite{bloesch2013state, bloesch2013state2, fink2020proprioceptive}, we do not rely on the assumptions of constant leg length ($\boldsymbol{\varrho}_t = \text{constant}$) or foot non-slippage ($\dot{s}_{i,t} = 0$). Instead, we explicitly consider the values of $\boldsymbol{\varrho}_t$ and $\dot{s}_{i,t} = 0$ in our measurement model \eqref{obs}.
\end{remark}

\begin{remark}
The propagation noise  \(\boldsymbol{n}_t\) in \eqref{process} and measurement noise \(\boldsymbol{\eta}_t\) in \eqref{obs} arise from various sources, such as sensors noise, modeling errors, and discretization errors. Therefore, it is generally impossible to acquire their exact distributions in practice. Following the convention in \cite{bloesch2013state2}, we assume that both the propagation and measurement noises follow  Gaussian distributions, i.e., \(\boldsymbol{n}_t \sim \mathcal{N}(0, Q_t)\) and \(\boldsymbol{\eta}_t \sim \mathcal{N}(0, \Sigma_t)\), where the covariance matrix \(Q_t\) and \(\Sigma_t\) serve as adjustable parameters, encapsulating the combined impact of the aforementioned sources of noise. 
\end{remark}

\section{Robust Dual Kalman filter}
\label{sec:method}
In the last section, we build a state space model for legged robot proprioceptive state estimation problem. In this section, we will give a formal definition of the state estimation problem and present two key challenges that hinder the applications of existing algorithms. Based on this, we will introduce the dual beta Kalman filter algorithm to tackle these two challenges.

\subsection{State Estimation Problem Formulation}
\label{sec:state_estimation}
The state estimation problem is to recursively estimate the robot's state $\boldsymbol{x}_t$ at each time step $t$, given the noisy sensor measurements, and control inputs. 
To better describe the estimation problem, we can adopt a probabilistic perspective, reformulating the SSM \eqref{model} as a hidden Markov model (HMM): 
\begin{subequations}
\label{eq.HMM}
\begin{align}
\boldsymbol{x}_t &\sim p(\boldsymbol{x}_t|\boldsymbol{x}_{t-1}, \boldsymbol{u}_{t-1}) = \mathcal{N}(\boldsymbol{x}_{t};f(\boldsymbol{x}_{t-1}, \boldsymbol{u}_{t-1}), Q_{t}), \label{eq.HMM:a} \\
\boldsymbol{y}_t &\sim p(\boldsymbol{y}_t|\boldsymbol{x}_{t}) = \mathcal{N}(\boldsymbol{y}_t; h(\boldsymbol{x}_t;\boldsymbol{\varrho}_t), \Sigma_t). \label{eq.HMM:b}
\end{align}
\end{subequations}
Here $p(\boldsymbol{x}_{t}|\boldsymbol{x}_{t-1}, \boldsymbol{u}_{t-1})$ represents the transition probability, and $p(\boldsymbol{y}_t|\boldsymbol{x}_t)$ is the measurement likelihood probability. Under this notation, the state estimation problem aims to determine the posterior probability density of the robot's state at every time step $t$, i.e. $p(\boldsymbol{x}_t|\boldsymbol{y}_{0:t}, \boldsymbol{u}_{0:t-1})$. To solve the state estimation problem, common KF-based estimation algorithms \cite{bloesch2013state, bloesch2013state2, teng2021legged} essentially perform a MAP estimation on the posterior probability density as
\begin{equation}
\label{MAP}
\begin{aligned}
\hat{\boldsymbol{x}}_{t} &= \mathop{\arg\max}\limits_{\boldsymbol{x}_t}p(\boldsymbol{x}_t|\boldsymbol{y}_{0:t},\boldsymbol{u}_{0:t-1}) \\
&= \mathop{\arg\max}\limits_{ \boldsymbol{x}_t}p(\boldsymbol{y}_t|\boldsymbol{x}_t)p(\boldsymbol{x}_t|\boldsymbol{y}_{0:t-1},\boldsymbol{u}_{0:t-1}) \\
&= \mathop{\arg\min}\limits_{\boldsymbol{x}_t} \big\{ -\log p(\boldsymbol{y}_t|\boldsymbol{x}_t) -\log p(\boldsymbol{x}_t|\boldsymbol{y}_{0:t-1},\boldsymbol{u}_{0:t-1})  \big\} \\
&=\mathop{\arg\min}\limits_{ \boldsymbol{x}_t} \underbrace{\big\{l_{h}^{\mathrm{CE}}(\boldsymbol{x}_t,\boldsymbol{y}_t) + l_f(\boldsymbol{x}_t, \boldsymbol{u}_{t-1})\big\}}_{\stackrel{\triangle}{=}\textstyle J_{\mathrm{CE}}},
\end{aligned}
\end{equation}
to obtain a point estimate $\hat{\boldsymbol{x}}_{t}$ of the robot's state at time $t$. Here, $l_h^{\mathrm{CE}}(\cdot,\cdot)$ denotes the measurement loss, and $l_f(\cdot, \cdot)$ represents the prior loss. 

The MAP problem \eqref{MAP} highlights the need for an accurate measurement likelihood. However, the current likelihood function $\mathcal{N}(\boldsymbol{y}_t; h(\boldsymbol{x}_t;\boldsymbol{\varrho}_t), \Sigma_t)$ in \eqref{eq.HMM:b} includes two unknowns—leg length parameters and foot slippage velocity. Under the influence of these two unknown variables, obtaining an accurate measurement function \eqref{obs} becomes infeasible. Consequently, the likelihood function cannot be precisely determined, thereby invalidating common KF-based algorithms that rely on solving the MAP problem \eqref{MAP}. To address these two unknowns, we propose a dual robust estimation framework.

\begin{figure}[!t]
\centering
\includegraphics[width=0.46\textwidth]{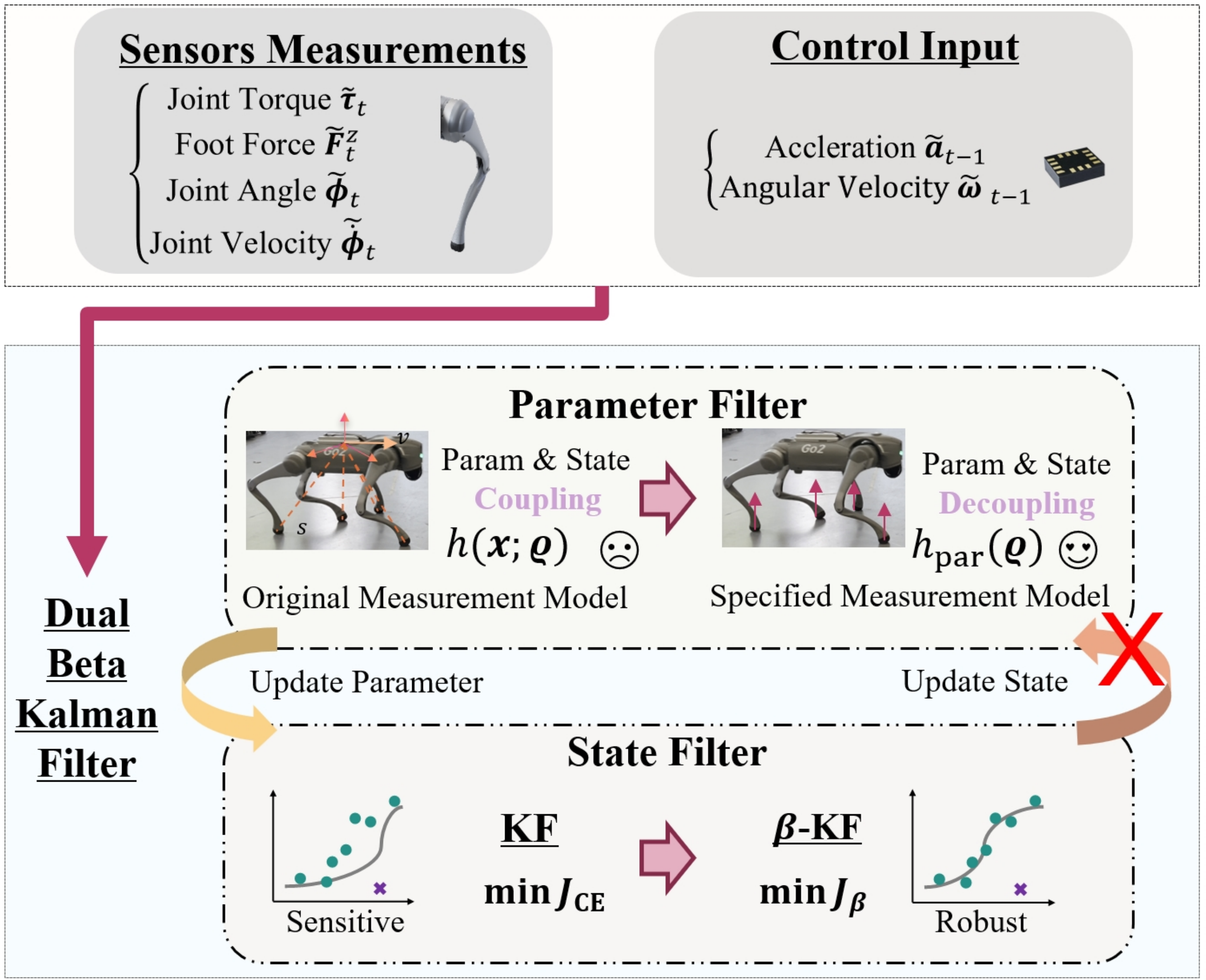}
\caption{Framework overview of the Dual $\beta$-KF. Proprioceptive sensor data are input to the Dual $\beta$-KF, where a state-independent parameter filter first estimates the leg length parameters. These estimates are then used by a robust state filter to determine the robot's state.}
\label{fig_3}
\end{figure}

\subsection{Dual Robust Estimation Framework}
\label{sec:dual_b}
The proposed dual robust estimation framework consists of two concurrently running filters, a parameter filter and a state filter, capable of simultaneously estimating leg length parameters and the robot's states, as illustrated in Fig.{\ref{fig_3}}.

For the parameter filter, its state propagation model is defined as a random walk process \cite{popovici2017dual, yang2022online}, 
\begin{equation}
\label{eq.param_model}
\boldsymbol{\varrho}_{t+1} = \boldsymbol{\varrho}_{t} + \boldsymbol{\xi}_{t},
\end{equation}
where $\boldsymbol{\xi}_{t}$ is Gaussian white noise. In the classical dual estimation framework \cite{popovici2017dual, wan2001dual, hong2014novel}, the measurement model of the parameter filter still uses \eqref{obs}, i.e, both parameters and states share the same measurement model. However, if the state estimate has significant errors, updating the parameters with it will amplify inaccuracies, leading to error accumulation in subsequent state updates. To avoid such an entanglement, we improve the classical dual estimation framework by introducing a separate measurement model for $\varrho_t$, decoupled from the states as 
\begin{equation}
\label{eq.param_meas}
\boldsymbol{z}_{t} = h_{\text{par}}(\boldsymbol{\varrho}_{t}) + \boldsymbol{\zeta}_t,
\end{equation}
where \( h_{\text{par}}(\cdot) \) represents the partial measurement function only for leg length parameters, which is independent of the robot states, while \(\boldsymbol{z}_t\) is the corresponding measurement vector (see \eqref{statics_meas} for details).
Note that, only proprioceptive sensor information is used in \(\boldsymbol{z}_t\) and \( g(\cdot) \), making our approach more practical compared to the method proposed in \cite{yang2022online}. The noise term \(\boldsymbol{\zeta}_t\) is also assumed to follow a Gaussian distribution.
This adjustment isolates the parameter updates from the influence of the states, allowing for more accurate parameter estimation. We will detail the design of this refined parameter filter in Sec.\ref{sec:param_filter}. 

For the state filter, its SSM is defined by \eqref{process} and \eqref{obs}. Unlike the varying leg length parameters, which can be estimated using the parameter filter introduced above, the current sensor configuration makes it impossible to directly measure the foot slippage velocity $\dot{\boldsymbol{s}}_{i,t}$.
Fortunately, foot slippage is a rare event, occurring in our preliminary tests with a probability of less than approximately 5\%. And when foot slippage does not occur, $\dot{\boldsymbol{s}}_{i,t}$ is effectively zero. Based on these facts, we can omit $\dot{\boldsymbol{s}}_{i,t}$ from \eqref{obs} and treat occurrences of foot slippage as outliers in the measurement likelihood \eqref{eq.HMM:b}. 
This approach enables us to address the issue by improving the robustness of the KF-based algorithms. 
The details of this modification will be presented in Sec.\ref{sec:state_filter}.


\subsection{Statics-based Parameters Filter}
\label{sec:param_filter}
The robot's leg length parameters are represented by $\boldsymbol{\varrho} = \left[l_{c,1}, \dots, l_{c,N} \right]^\top \in \mathbb{R}^N$, where $l_{c,i}$ denotes the length of the $i$-th calf leg. This representation accounts for the fact that the calf leg length is primarily affected by dynamic deformation during locomotion \cite{bloesch2013kinematic, yang2022online}. We aim to identify a measurement model for the parameter filter that excludes state variables and relies solely on proprioceptive sensor information. In this context, a natural candidate for such a measurement model is the leg dynamic equation \cite{craig2005introduction}, as it incorporates the leg length parameters without including the robot's state, and depends exclusively on three types of proprioceptive sensors: foot contact force sensors, joint encoders, and joint torque sensors. This equation is given by
\begin{equation}\label{dynamic}
\begin{aligned}
\boldsymbol{M}(\boldsymbol{\phi}_i)\ddot {\boldsymbol{\phi}}_i + \boldsymbol{V}(\boldsymbol{\phi}_i, \dot {\boldsymbol{\phi}}_i) - \boldsymbol{J}(\boldsymbol{\phi}_i; l_{c,i})^\top \boldsymbol{F}_{i} = \boldsymbol{\tau}_i ,
\end{aligned}
\end{equation}
where $\boldsymbol{M}(\boldsymbol{\phi}_i) \in 
\mathbb{R}^{3\times 3}$ is the inertia matrix, $\boldsymbol{V}(\boldsymbol{\phi}_i, \dot {\boldsymbol{\phi}}_i) \in 
\mathbb{R}^{3}$ is the vector of centrifugal, Coriolis, and gravity torques, $\boldsymbol{F}_{i} \in 
\mathbb{R}^{3}$ is the Ground Reaction Force (GRF) of the $i$-th leg  and $\boldsymbol{\tau}_i \in \mathbb{R}^{3}$ is the vector of all joint torques of the $i$-th leg . However, the joint angular acceleration $\ddot {\boldsymbol{\phi}}_i$ is typically too noisy to be usable \cite{kang2023external}, which means that \eqref{dynamic} cannot be directly used as the parameter filter measurement model. Luckily, as illustrated in \cite{kang2023external}, \eqref{dynamic} can be approximated using the leg statics equation: 
\begin{equation}
\label{statics}
\boldsymbol{\tau}_i = - \boldsymbol{J}(\boldsymbol{\phi}_i; l_{c,i})^{\top} \boldsymbol{F}_{i},
\end{equation}
since the mass of the leg is relatively small, and when the foot is in contact with the ground, the change in leg velocity is minimal.
Compared to the leg dynamic equation \eqref{dynamic}, the leg static equation \eqref{statics} no longer includes $\ddot{\boldsymbol{\phi}}_i$ and is computationally simpler.
Unfortunately, using \eqref{statics} directly as the parameter filter measurement model is intractable because foot contact force sensors can only capture the normal component of the GRF, $\boldsymbol{F}_{i}^{z}$, which corresponds to the last component of the GRF. This means that the information required for parameter estimation using \eqref{statics} is incomplete.
Nonetheless, since $\boldsymbol{J}(\boldsymbol{\phi}_i; l_{c,i})$ is invertible for a leg with three joints \cite{fink2020proprioceptive, menner2024simultaneous}, we can perform a simple transformation on \eqref{statics} to convert it into a tractable parameter filter measurement model. Specifically, by multiplying both sides of \eqref{statics} by $\boldsymbol{J}(\tilde {\boldsymbol{\phi}}_{1,t}; l_{c,1, t})^{-\top}$, then taking the last component of both sides, we obtain
\begin{equation}
\label{statics_meas}
\begin{aligned}
\begin{bmatrix} \tilde {\boldsymbol{F}}_{1,t}^{z} \\
...\\
\tilde {\boldsymbol{F}}_{N,t}^{z}
\end{bmatrix} &= -\boldsymbol{\lambda} \begin{bmatrix} \boldsymbol{J}(\tilde {\boldsymbol{\phi}}_{1,t}; l_{c,1, t})^{-\top}\tilde {\boldsymbol{\tau}}_{1, t} \\
...\\
\boldsymbol{J}(\tilde {\boldsymbol{\phi}}_{N,t}; l_{c,N, t})^{-\top}\tilde {\boldsymbol{\tau}}_{N, t}
\end{bmatrix}  + \boldsymbol{\zeta}_{t},  
\end{aligned}
\end{equation}
where $\boldsymbol{\lambda} = \underbrace{[0, 0, 1, \ldots, 0, 0, 1]}_{N \times}$; $\tilde {\boldsymbol{F}}_{i,t}^{z}$ and $\tilde {\boldsymbol{\tau}}_{i, t}$ are noisy measurements that can be directly obtained from the sensors, making \eqref{statics_meas}  tractable. The propagation model \eqref{eq.param_model} and the specified measurement model \eqref{statics_meas} together constitute the statics-based parameter filter's SSM. By applying typical filtering algorithms, such as the EKF or UKF, on this SSM, we can obtain leg length parameter estimates, $\hat{\boldsymbol{\varrho}}_{t}$, at each time step $t$.

\subsection{\texorpdfstring{State filter: $\beta$-Kalman filter}{Beta-Kalman filter}}
\label{sec:state_filter}
In Sec.\ref{sec:dual_b}, we discussed that the occurrences of foot slippage can be treated as measurement outliers. However, common KF-based algorithms cannot handle outliers \cite{cao2022robust}, as their objective is to solve the MAP problem \eqref{MAP}, where $l_h^{\mathrm{CE}}(\boldsymbol{x}_t, \boldsymbol{y}_t)$ essentially represents a KL divergence, $\mathcal{D}_{\text{KL}}(p^t_{\text{true}}(\boldsymbol{y}_t)||p(\boldsymbol{y}_t|\boldsymbol{x}_t))$, between the true measurement distribution $p^t_{\text{true}}(\boldsymbol{y}_t)$ and measurement likelihood $p(\boldsymbol{y}_t|\boldsymbol{x}_t)$, which is highly sensitive to outliers \cite{futami2018variational}.

Inspired by robust Bayesian inference theory \cite{knoblauch2022optimization}, to enhance the robustness of KF-based algorithms, we replace the KL divergence with a robust divergence, $\beta$-divergence, which has a truncation effect on the influence of outliers \cite{futami2018variational, knoblauch2022optimization}, and is calculated as 
\begin{equation}\label{beta_divergence} 
\begin{aligned}
&\mathcal{D}_{\beta}\left[p^t_{\text{true}}(\boldsymbol{y}_t) || p(\boldsymbol{y}_t|\boldsymbol{x}_t)\right]  = \frac1\beta\int p^t_{\text{true}}(\boldsymbol{y}_t)^{1+\beta}\text{d} \boldsymbol{y}_t \\
&-\frac{\beta+1}{\beta}\int p^t_{\text{true}}(\boldsymbol{y}_t)p(\boldsymbol{y}_t|\boldsymbol{x}_t)^{\beta}\text{d} \boldsymbol{y}_t+\int p(\boldsymbol{y}_t|\boldsymbol{x}_t)^{1+\beta}\text{d} \boldsymbol{y}_t.    
\end{aligned}
\end{equation}
Here, we usually take $p^t_{\text{true}}(\boldsymbol{y}_t)$ as the empirical distribution of the true measurement data, i.e., $p^t_{\text{true}}(\boldsymbol{y}_t) = \delta(\boldsymbol{y} - \boldsymbol{y}_t)$, where $\delta(\cdot)$ is a Dirac delta function. We can remove the first term in \eqref{beta_divergence} because it is independent of the optimization variable $\boldsymbol{x}_t$ and substitute $p^t_{\text{true}}(\boldsymbol{y}_t)$, then minimizing \eqref{beta_divergence} is equivalent to minimizing the following $l_{h}^{\beta}(\boldsymbol{x}_t,\boldsymbol{y}_t)$ \cite{cao2022robust}, which is given by 
\begin{equation}\label{beta_2}
\begin{aligned}
l_{h}^{\beta}(\boldsymbol{x}_t,\boldsymbol{y}_t) = -\frac{\beta+1}{\beta} p(\boldsymbol{y}_t|\boldsymbol{x}_t)^{\beta} + \int p(\boldsymbol{y}_t|\boldsymbol{x}_t)^{1+\beta}\text{d} \boldsymbol{y}_t.
\end{aligned}
\end{equation}
We denote $\left\| z \right\|^2_{A} = z^TAz$, and substitute \eqref{eq.HMM:b} and the leg length parameter estimates $\hat {\boldsymbol{\varrho}}_{t}$ into  \eqref{beta_2}, we have
\begin{equation}\label{l_beta}
\begin{aligned}
l_{h}^{\beta}(\boldsymbol{x}_t,\boldsymbol{y}_t) &= -\frac{(\beta+1)\exp \{\frac{\beta}{2}\left\| \boldsymbol{y}_t - h(\boldsymbol{x}_t; \hat{\boldsymbol{\varrho}}_t) \right\|^2_{\Sigma_t^{-1}} \}}{\beta(2\pi)^{\frac{\beta m}{2}}\det(\Sigma_t)^{\frac{\beta}{2}}} \\
&+ \frac{1}{(\beta + 1)^{\frac{1}{2}}(2\pi)^{\frac{(\beta-2) m}{2}}\det(\Sigma_t)^{\frac{\beta - 2}{2}}}.
\end{aligned}
\end{equation}
The prior loss in the MAP problem \eqref{MAP} can be computed using \eqref{eq.HMM:a}, we drop the irrelevant terms and denote the result as
\begin{equation}\label{l_f}
\begin{aligned}
l_f(\boldsymbol{x}_t,\boldsymbol{u}_{t-1}) &= \frac{1}{2}\left\| \boldsymbol{x}_t - f(\hat {\boldsymbol{x}}_{t-1}, \boldsymbol{u}_{t-1}) \right\|^2_{P_{t|{t-1}}^{-1}},
\end{aligned}
\end{equation}
where $P_{t|t-1}$ is the prior error covariance matrix at time $t$, which can be derived by the Riccati equation:
\begin{equation}
\label{eq.pre_riccati}
\begin{aligned}
P_{t+1|t} &= Q_t + F_tP_{t|t-1}F_t^T \\
&- F_tP_{t|t-1}H_t^T(H_tP_{t|t-1}H_t^T + R_t)^{-1}H_tP_{t|t-1}F_t^T,
\end{aligned}
\end{equation}
where $F_t = \frac{\partial f}{\partial \boldsymbol{x}_t}$ and $H_t = \frac{\partial h}{\partial \boldsymbol{x}_t}$ are the transition Jacobian matrix and the measurement Jacobian matrix, respectively. In summary, the final optimization objective is transformed from \eqref{MAP} into a more robust form as
\begin{equation}
\label{eq.final}
\begin{aligned}
\hat{\boldsymbol{x}}_t
= \mathop{\arg\min}\limits_{ \boldsymbol{x}_t} \underbrace{\{l_{h}^{\beta}(\boldsymbol{x}_t,\boldsymbol{y}_t) + l_f(\boldsymbol{x}_t, \boldsymbol{u}_{t-1})\}}_{\stackrel{\triangle}{=}\textstyle J_{\beta}}.
\end{aligned}
\end{equation}
The $\beta$-KF involves solving the unconstrained optimization problem \eqref{eq.final} at each time step $t$ to obtain the robot state estimate $\hat{\boldsymbol{x}}_t$. Notably, the hyperparameter $\beta \in (0,1)$ determines the robustness of the $\beta$-KF against outliers, with the chosen value of $\beta$ being positively correlated with the severity of the outliers. When $\beta$ approaches 0, the $\beta$-divergence converges to the KL divergence, and the $\beta$-KF degenerates into common KF-based algorithms.
\begin{remark}
The mathematical theory behind $\beta$-divergence is highly intricate. Previous studies have shown that the influence function of $\beta$-divergence is bounded \cite{cao2022robust}, making it less sensitive to outliers. However, providing theoretical proof for the enhanced robustness of $\beta$-KF is highly challenging and non-trivial. This theoretical exploration will be addressed in our future work.
\end{remark}
\begin{remark}
The statics-based parameter filter measurement model \eqref{statics_meas} is unaffected by the unknown foot slippage velocity. As a result, \eqref{statics_meas} does not introduce significant outliers, making a basic EKF or UKF sufficient for the parameter filter. Here, we choose the UKF \cite{bloesch2013state2} as the parameter filter, and the pseudocode for the proposed Dual $\beta$-KF is provided in Algorithm \ref{alg.1}.
\end{remark}

\begin{algorithm}
\caption{Dual $\beta$-KF}\label{alg.1}
\renewcommand{\algorithmicrequire}{\textbf{Input:}}
\renewcommand{\algorithmicensure}{\textbf{Output:}}
\begin{algorithmic}[1]

\REQUIRE Initial state estimate $\hat {\boldsymbol{x}}_{0}$ and initial parameter estimate  $\hat {\boldsymbol{\varrho}}_{0}$, initial covariance $P_{0}$, hyperparameter $\beta$

\ENSURE State estimates $\hat {\boldsymbol{x}}_{1:T}$, parameter estimates $\hat {\boldsymbol{\varrho}}_{1:T}$
\STATE $P_{1|0} \leftarrow F_{0}P_0F_0^\top + Q_0$
\FOR{each time step $t$}
    \STATE Obtain the  proprioceptive sensor data $\tilde{\boldsymbol{a}}_{t-1}$, $\tilde{\boldsymbol{\omega}}_{t-1}$, $\tilde {\boldsymbol{\phi}}_t$, $\tilde {\dot {\boldsymbol{\phi}}}_t$, $\tilde {\boldsymbol{\tau}}_t$, $\tilde {\boldsymbol{F}}^{z}_t$
    \STATE \textbf{Perform parameter filter:}
        \STATE Based on \eqref{eq.param_model}, execute the UKF predict step  
        \STATE Based on \eqref{statics_meas}, execute the UKF update step to obtain $\hat{\boldsymbol{\varrho}}_{t}$
    \STATE \textbf{Perform state filter:}
    \STATE Substituting  $\tilde{\boldsymbol{a}}_{t-1}$, $\tilde{\boldsymbol{\omega}}_{t-1}$ and $\hat {\boldsymbol{x}}_{t-1}$ into \eqref{process} yields $f(\hat {\boldsymbol{x}}_{t-1}, \boldsymbol{u}_{t-1})$
    \STATE Substituting $\hat {\boldsymbol{\varrho}}_{t}$ into \eqref{obs} yields $h(\boldsymbol{x}_t; \hat {\boldsymbol{\varrho}}_{t})$
    \STATE Solve the optimization problem \eqref{eq.final} to get $\hat {\boldsymbol{x}}_{t}$
    \STATE Calculate  $P_{t+1|t}$ through  \eqref{eq.pre_riccati}
\ENDFOR
\end{algorithmic}
\end{algorithm}

\vspace{-3mm}
\section{Experiments}
\label{sec:exp}
We conduct a series of experiments to demonstrate the performance of our proposed algorithms. We first test our algorithms in the Gazebo simulation, 
and second, conducted the real-world experiment on a Unitree GO2 robot. 

To validate the effectiveness of our proposed \textbf{Dual} $\boldsymbol{\beta}$\textbf{-KF}, we compare its experimental results with those of the following algorithms: (1) \textbf{QEKF} \cite{bloesch2013state, fink2020proprioceptive}, quaternion-based EKF; and (2) \textbf{UKF-OR} \cite{bloesch2013state2}, UKF with outlier rejection. These two algorithms represent the most commonly used state-of-the-art proprioceptive state estimation techniques for legged robots. 
In addition, we include (3) \textbf{Dual QEKF}, which employs the proposed dual estimation framework with statics-based parameter filter with a statics-based parameter filter to estimate the varying leg length while continuing to utilize QEKF as the state filter, and (4) $\boldsymbol{\beta}$\textbf{-KF}, where only the proposed $\beta$-KF is applied while still using a fixed leg length, as setups for the ablation study. These setups are designed to illustrate the effectiveness of the proposed algorithm components.

We implemented all the above algorithms on Ubuntu 22.04 using Python 3.8 and employed CasADi as the solver for nonlinear optimization problems in the $\beta$-KF.
\vspace{-2mm}
\subsection{Evaluation Metrics}
\label{eval}
To quantitatively evaluate the estimation accuracy, we define the following metrics, as referenced in \cite{yang2023multi} and \cite{kim2021legged}:
\begin{itemize}
\item{
\textbf{ATE} 
 (m): $\sqrt{\frac{1}{T}\Sigma_{t=0}^T\left\| \boldsymbol{p}_t - \hat {\boldsymbol{p}}_t \right\|^2}$

 The absolute translation error (ATE) measures the global consistency error between the estimated trajectory and the ground truth trajectory. 
}
\item{
\textbf{MPD} (m): $\max_{t=0:T} \left\| \boldsymbol{p}_t - \hat {\boldsymbol{p}}_t \right\|$

The maximum position drift (MPD) measures the largest absolute estimation deviation over the entire trajectory traveled.
}
\item{
\textbf{DR} (\%): $\frac{\left\| \boldsymbol{p}_T - \hat {\boldsymbol{p}}_T \right\|}{l_{\text{traj}}}$

The final position drift ratio (DR) measures the endpoint position error relative to the traveled distance, indicating the ultimate drift in the estimated trajectory.
}
\end{itemize}
Here, $\boldsymbol{p}$ indicates the ground truth body position, $\hat {\boldsymbol{p}}$ indicates the estimated position, $l_{\text{traj}}$ is the motion trajectory length, and $T$ is the total duration of motion.

\vspace{-2mm}
\subsection{Experimental Setup}

\begin{figure}[!t]
\centering
\subfloat{
    \includegraphics[width=0.22\textwidth, height=0.15\textwidth]{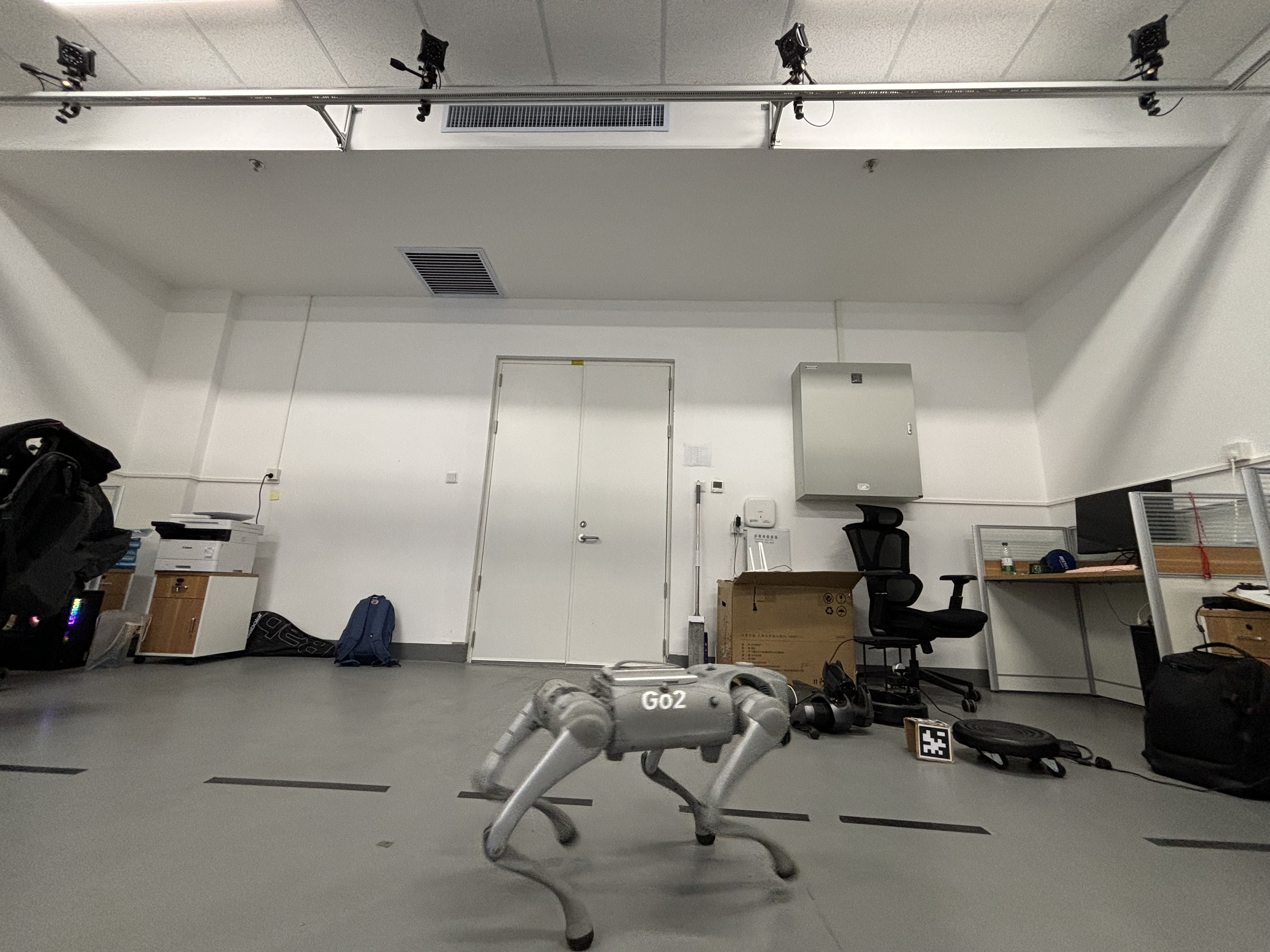}
    \label{fig4:sim_robot}
}
\hfill
\subfloat{
    \includegraphics[width=0.22\textwidth, height=0.15\textwidth]{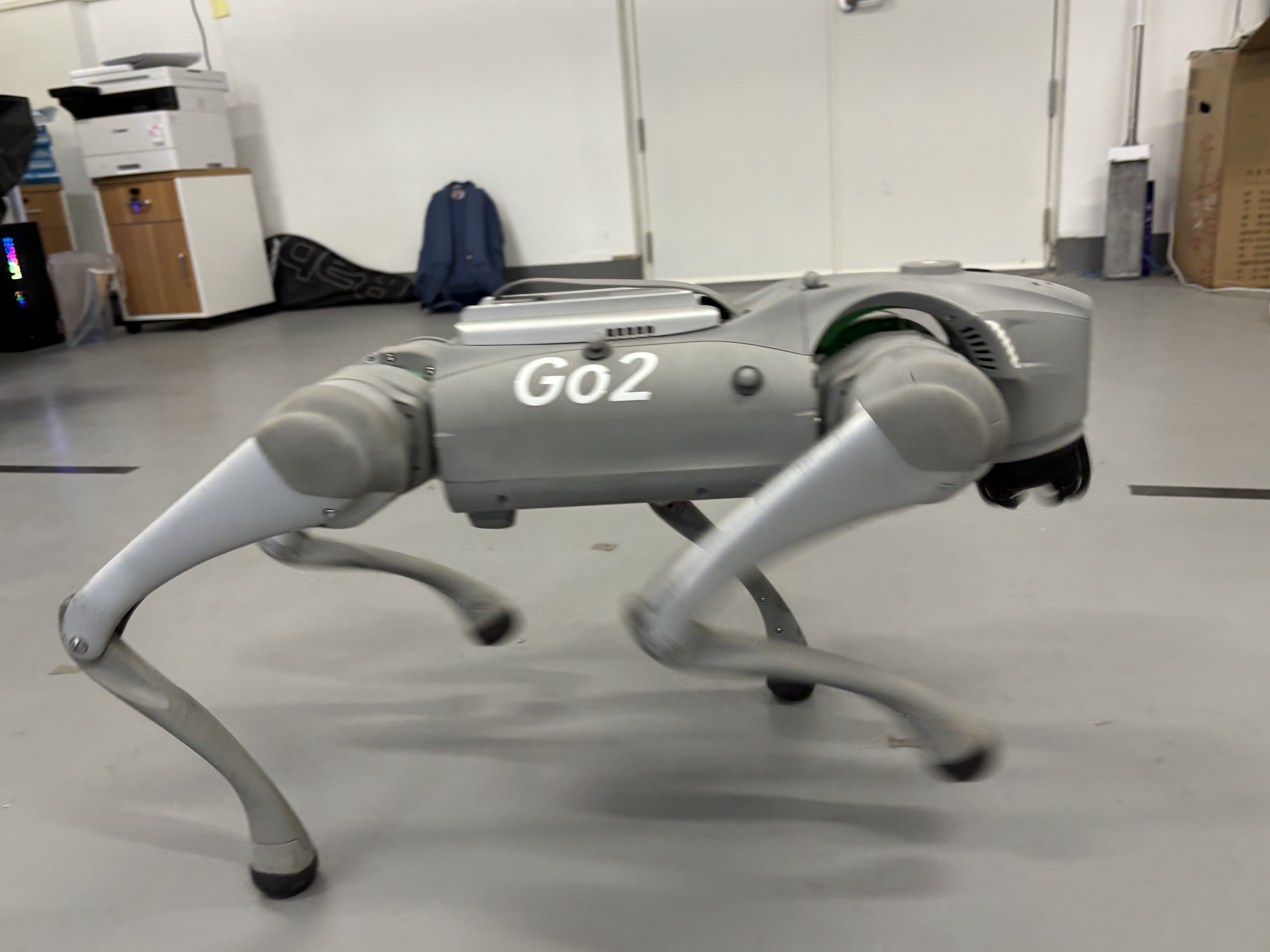}
    \label{fig4:real_robot}
}

\caption{A Unitree Go2 robot walks with a trot gait in a laboratory equipped with motion capture system.}
\label{fig_4}
\end{figure}

In all experiments, the legged robot moves in a trot gait on flat ground, and sensor data is collected during the movement, including one IMU, 12 joint encoders, joint torque sensors, and four foot contact force sensors. All proprioceptive sensor data is recorded at 500 Hz. Additionally, the robot's ground-truth pose is obtained from the Gazebo environment in simulation and provided by a high-precision motion capture system in real-world experiments.

Approximately 2 minutes of motion data are collected in the Gazebo simulation, and 1 minute of data is collected in the real-world experiment. All algorithms are run offline on the collected data using an Intel Core i9-12900K CPU. The initial state for all estimation algorithms is set to the ground truth, the initial leg length parameters are set to the measured values and IMU biases are initialized to zero. The tunable hyperparameter $\beta$ is set to $3 \times 10^{-6}$ for the Gazebo simulation and $1 \times 10^{-3}$ for real-world experiments. This configuration is based on the observation that foot slippage occurs more frequently in real-world experiments, resulting in more severe outliers. Consequently, a larger $\beta$ is required for real-world experiments, as explained in Sec.\ref{sec:state_estimation}.

\begin{figure*}[!t]
\centering
\subfloat{
    \includegraphics[width=0.40\textwidth]{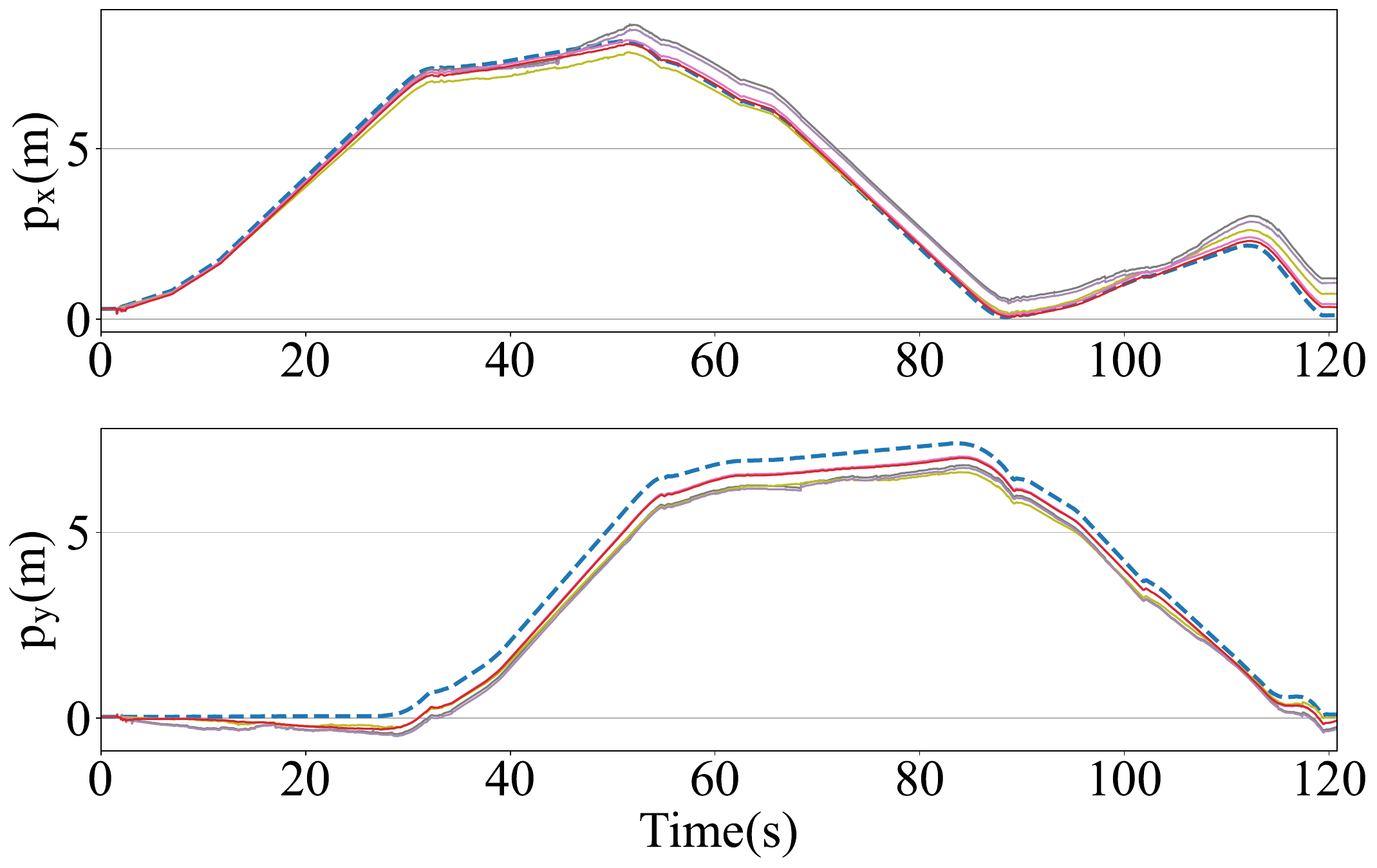}
    \label{fig5:sim}
}
\subfloat{
    \includegraphics[width=0.40\textwidth]{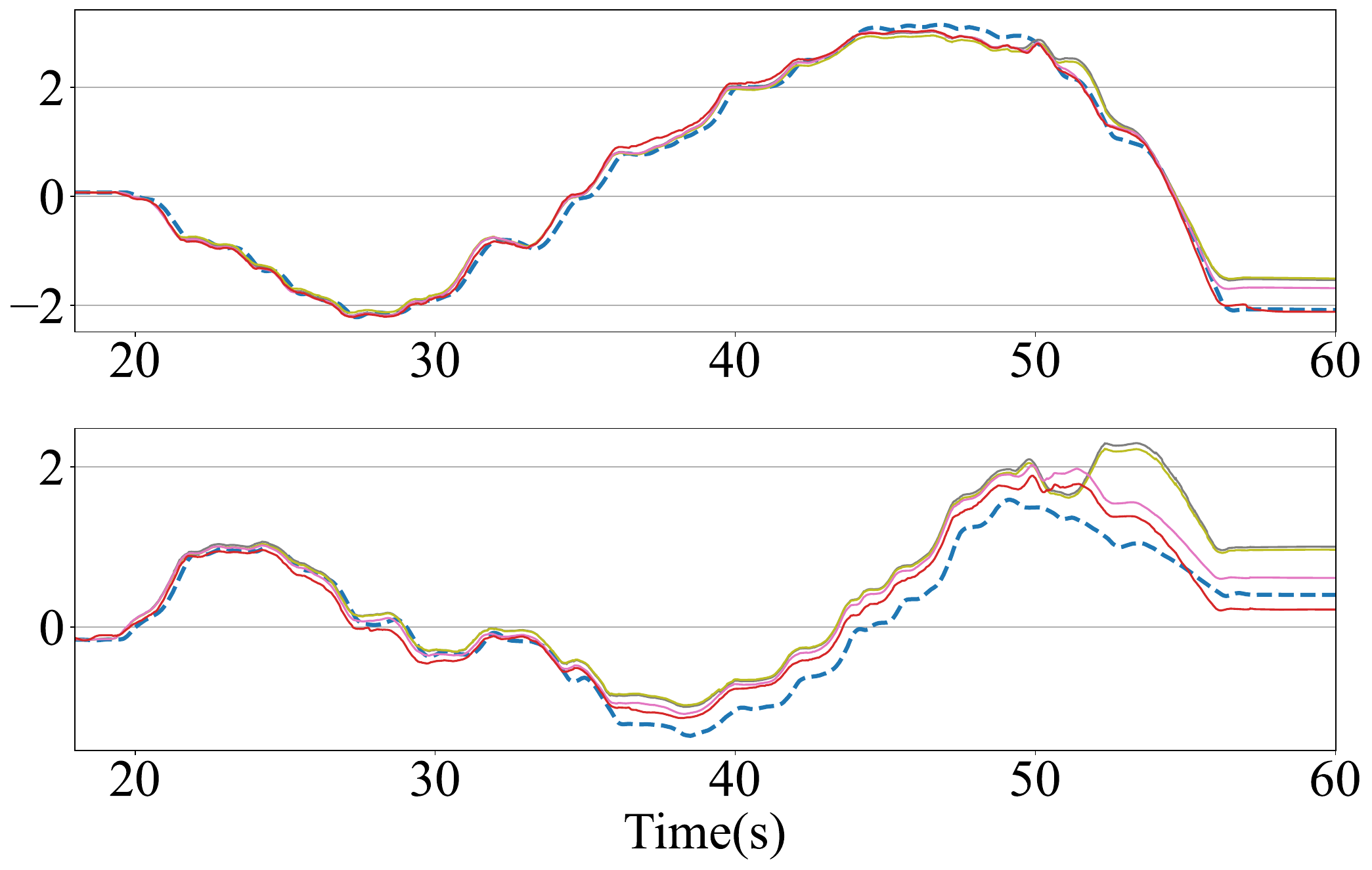}
    \label{fig5:real}
}
 \subfloat{
    \includegraphics[width=0.145\textwidth]{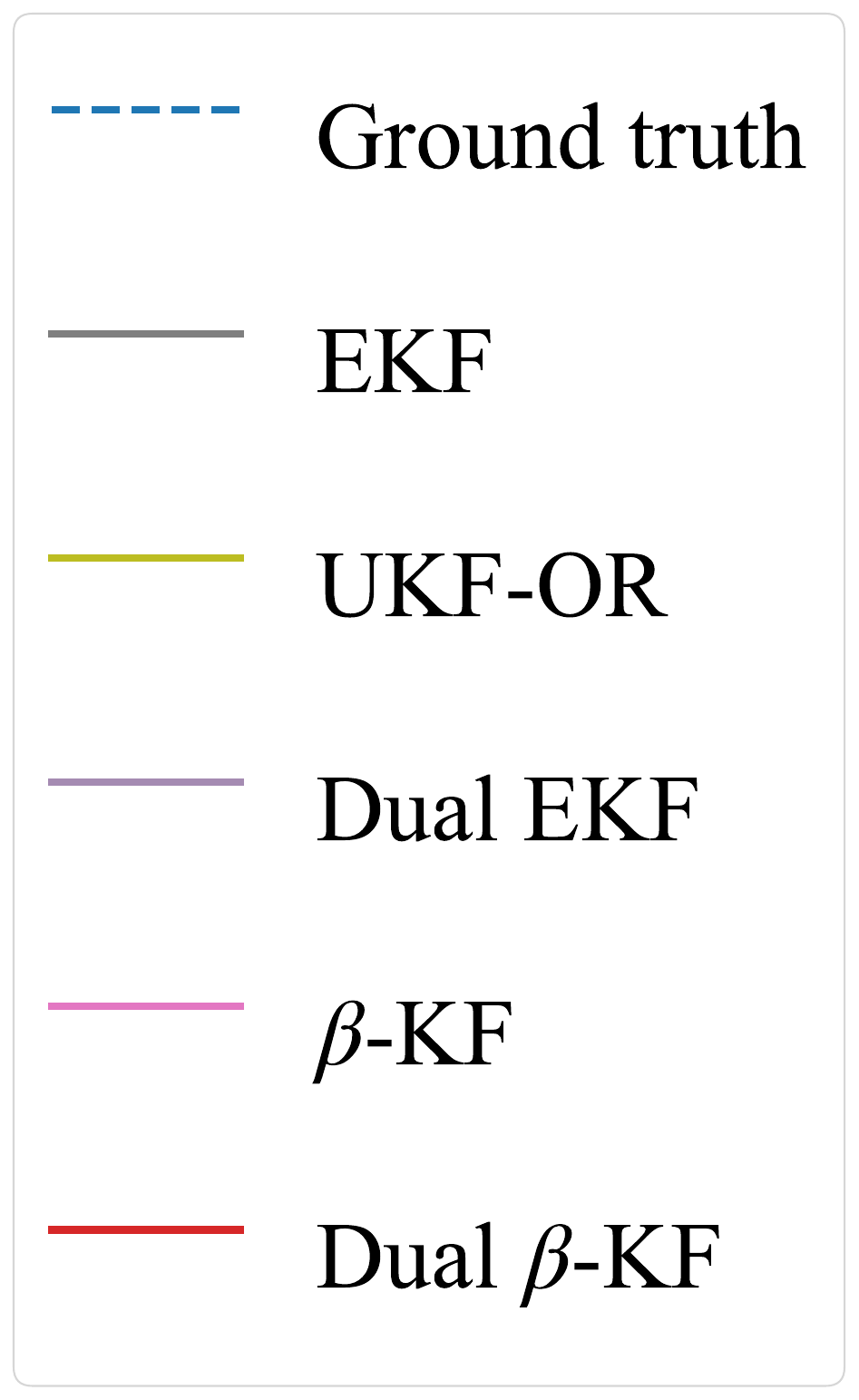}
        \label{fig5:legend}
    \phantomcaption
    }
\caption{Position estimation results for different algorithms in the simulation (left) and the real-world experiment (right) using the Unitree GO2 robot.}
\label{fig5:result}
\end{figure*}



\begin{figure}[!t]
\centering
\subfloat{
    \includegraphics[width=0.45\textwidth]{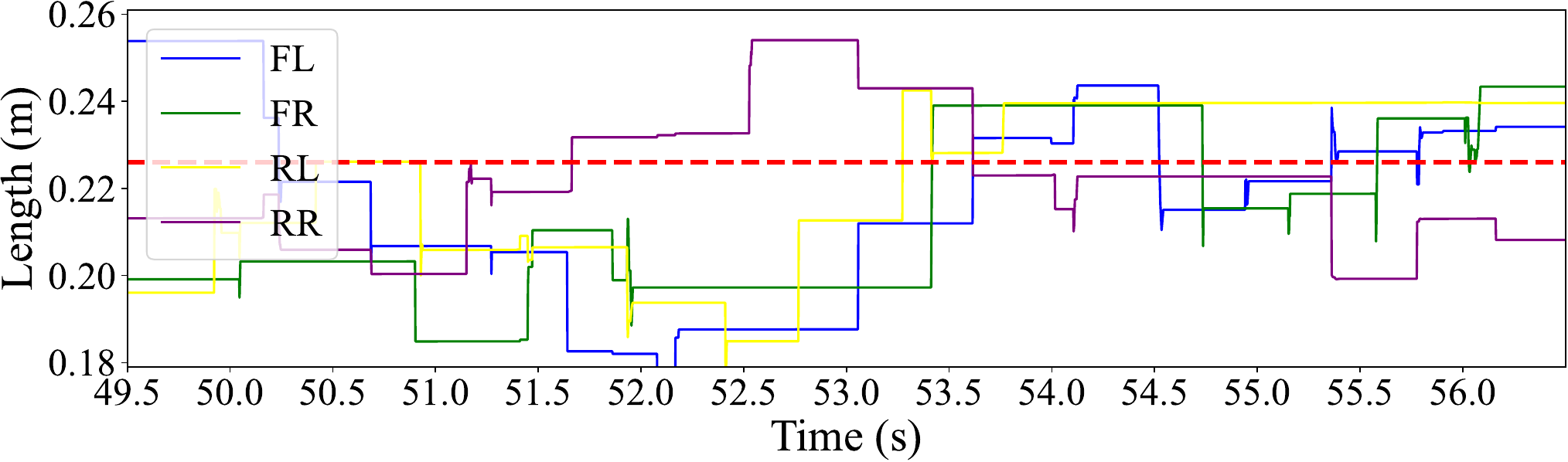}
    \label{fig6:calib_lc}
}

\subfloat{
    \includegraphics[width=0.45\textwidth]{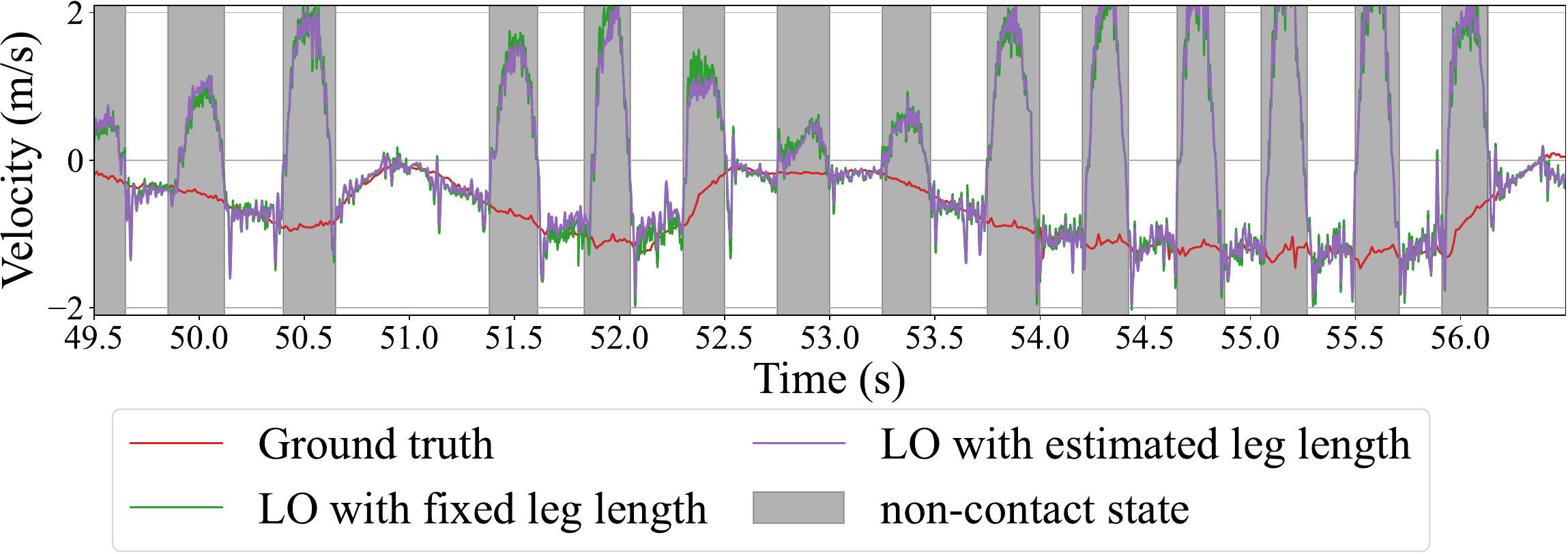}
    \label{fig6:calib_lo}
}
\caption{Above: The estimated length of each calf leg obtained from the parameter filter, with the red dashed line representing the measured leg length, which is 0.226m. FL(front-left), FR(front-right), RL(rear-left), and RR(rear-right) correspond to the four legs of the robot. Below: LO velocity of the FR leg: fixed leg length v.s. estimated leg length. The black shaded areas have the same meaning as in Fig.\ref{fig1.c}.}
\label{fig7}
\end{figure}

\vspace{-2mm}
\subsection{Comparison Results}                   

Table \ref{table:results}
 presents the results of different estimation algorithms. As expected, the baseline QEKF exhibits the largest estimation error due to its lack of handling for an inaccurate measurement model. Conversely, the UKF-OR can effectively suppress measurement outliers by fine-tuning the threshold, thereby improving estimation accuracy. As shown on the left side of Fig. \ref{fig5:result}, the QEKF shows a spike around 100s in the x-direction position estimate, which is largely eliminated after applying the UKF-OR. However, because the UKF-OR disregards measurement updates when the filter innovation exceeds the threshold, it fails to utilize some useful information for state updates, causing larger errors at certain times, as shown in Fig.\ref{fig5:result}.
 In contrast, the proposed $\beta$-KF automatically assigns low weights to outliers, allowing for better utilization of measurement information. As a result, it significantly improves estimation accuracy compared to the UKF-OR. 

The leg length parameter estimation results from the dual estimation framework in the real-world experiment are shown at the top of Fig.\ref{fig7}, with a data segment from 49.5s to 56.5s highlighted for clarity. Compared to a fixed leg length of 0.226m, utilizing leg length estimates ranging from 0.182m to 0.253m reduces LO velocity calculation error, as shown in Fig.\ref{fig7}. This demonstrates that the simultaneous estimation of leg length parameters improves the accuracy of the measurement model \eqref{obs}.
Table \ref{table:results} and Fig.\ref{fig5:result} clearly show that the estimator with the dual estimation framework, benefiting from a more accurate measurement model, outperforms the one without the dual estimation framework (i.e., using a fixed leg length) in the real-world experiment. 
In simulations, the dual estimation framework has limited impact as leg length in Gazebo remains nearly constant, whereas in real-world settings, dynamic changes in leg length due to deformation make it significantly more effective.
This discrepancy highlights the practical relevance of our method and supports its correctness. Notably, the Dual $\beta$-KF achieves the best estimation accuracy, providing over a 40\% improvement compared to the baseline QEKF. 

\begin{table}[t]
\scriptsize 
\caption{{Experiments Results on different environs}}\vspace{-2mm} 
\label{table:results}
\centering
\resizebox{0.48\textwidth}{!}{
\begin{tabular}{*{9}{c}}\toprule
\multirow{1}{*}{} & \multicolumn{3}{c}{\multirow{1}{*}{{Gazebo simulation ($\approx$ 35m)}}} 
  & \multicolumn{3}{c}{\multirow{1}{*}{{Real world ($\approx$ 20m)}}}\\ \midrule
{} & 
\multicolumn{1}{c}{{ATE (m)}} &
\multicolumn{1}{c}{{MPD (m)}} &
\multicolumn{1}{c}{{DR}} & 
\multicolumn{1}{c}{{ATE (m)}} &
\multicolumn{1}{c}{{MPD (m)}} &
\multicolumn{1}{c}{{DR}} \\ 
\midrule
\multirow{1}{*}{QEKF} & 
\multirow{1}{*}{0.714}  & \multirow{1}{*}{1.212} & \multirow{1}{*}{3.64\%} & 
\multirow{1}{*}{{0.412}} & \multirow{1}{*}{{0.121}} &   \multirow{1}{*}{7.12\%} \\ 

\multirow{1}{*}{{UKF-OR}} & 
\multirow{1}{*}{0.607} & \multirow{1}{*}{0.905} & \multirow{1}{*}{2.12\%} & 
\multirow{1}{*}{{0.410}} &  \multirow{1}{*}{1.169} &  \multirow{1}{*}{6.38\%} \\ 

\multirow{1}{*}{{Dual QEKF}} & 
\multirow{1}{*}{0.717} &  \multirow{1}{*}{1.089} & \multirow{1}{*}{3.32\%} & 
\multirow{1}{*}{{0.270}}  & \multirow{1}{*}{0.690} &  \multirow{1}{*}{3.77\%} \\ 

\multirow{1}{*}{{$\beta$-KF}} & 
\multirow{1}{*}{$\textbf{0.390}$}  & \multirow{1}{*}{$\textbf{0.552}$} & \multirow{1}{*}{1.24\%} & 
\multirow{1}{*}{{0.292}}  & \multirow{1}{*}{0.665} &  \multirow{1}{*}{3.63\%} \\ 

\rowcolor{tabGray}\multirow{1}{*}{{$\textbf{Dual $\beta$-KF}$}} & 
\multirow{1}{*}{0.400}  & \multirow{1}{*}{0.585} & \multirow{1}{*}{$\textbf{1.05}$\%} & 
\multirow{1}{*}{$\textbf{0.240}$}  & \multirow{1}{*}{$\textbf{0.499}$} &  \multirow{1}{*}{$\textbf{2.72}$\%} \\
\bottomrule
\end{tabular}}
\end{table}




\vspace{-2mm}
\section{Conclusion}
\label{sec:conclusion}
This paper proposes Dual $\beta$-KF, a robust state estimator for legged robots that relies solely on proprioceptive sensors. By introducing a measurement model that accounts for both foot slippage velocity and changes in leg length, we designed a dual robust estimation framework. This framework includes a statics-based parameter filter capable of estimating the leg length parameter independently of state variables, and a state filter, $\beta$-KF, which is robust to outliers caused by foot slippage, for estimating the robot's state. Experimental results show that our proposed Dual $\beta$-KF achieves significant improvements over previously used proprioceptive state estimation algorithms for legged robots.

\bibliographystyle{IEEEtran}
\bibliography{ref}


\end{document}